\documentclass[a4paper, 10pt, conference]{IEEEtran}  

\IEEEoverridecommandlockouts                              

\usepackage{hyperref}
\usepackage{graphicx}
\usepackage{cite}
\usepackage{amsmath,amssymb,amsfonts}
\usepackage{algorithmic}
\usepackage{textcomp}
\usepackage{xcolor}
\usepackage{multirow}
\usepackage{dblfloatfix}
\def\BibTeX{{\rm B\kern-.05em{\sc i\kern-.025em b}\kern-.08em
    T\kern-.1667em\lower.7ex\hbox{E}\kern-.125emX}}
\graphicspath{ {./figures/} }

\title{
SwipeBot: DNN-based Autonomous Robot Navigation among Movable Obstacles in Cluttered Environments
}

\author{\IEEEauthorblockN{
Nikolay Zherdev, Mikhail Kurenkov, Kristina Belikova and Dzmitry Tsetserukou
}
\IEEEauthorblockA{\textit{ISR Laboratory, Skolkovo Institute of Science and Technology, Moscow, Russia}}
\IEEEauthorblockA{$\{$Nikolay.Zherdev, Mikhail.Kurenkov, Kristina.Belikova, D. Tsetserukou$\}$@skoltech.ru}}

\begin{document}

\maketitle
\thispagestyle{empty}
\pagestyle{empty}

\begin{abstract}
In this paper, we propose a novel approach to wheeled robot navigation through an environment with movable obstacles. A robot exploits knowledge about different obstacle classes and selects the minimally invasive action to perform to clear the path. We trained a convolutional neural network (CNN), so the robot can classify an RGB-D image and decide whether to push a blocking object and which force to apply.  After known objects are segmented, they are being projected to a cost-map, and a robot calculates an optimal path to the goal. If the blocking objects are allowed to be moved, a robot drives through them while pushing them away. We implemented our algorithm in ROS, and an extensive set of simulations showed that the robot successfully overcomes the blocked regions.
Our approach allows a robot to successfully build a path through regions, where it would have stuck with traditional path-planning techniques.

\end{abstract}


\section{Introduction}

\begin{figure}[!t]
    \centering
    \includegraphics[width=8cm]{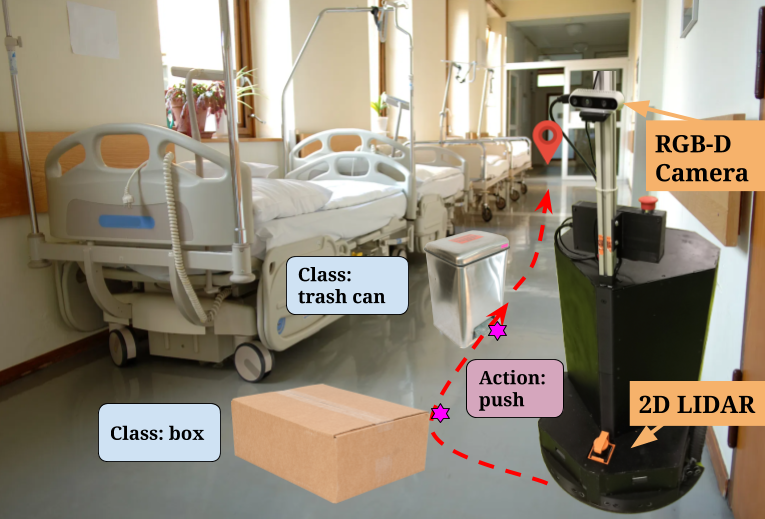}
    \caption{SwipeBot robot in cluttered environment. The single path is blocked by movable obstacles, which a robot can push away to continue its mission.}
    \label{fig:Thesis_illustration}
    \vspace{-0.5em}
\end{figure}

Operating in a cluttered environment is still an open problem for indoor mobile robots. For safe navigation, robots build and periodically rebuild a path to the goal, which is collision-free from static and dynamic obstacles. However, in case the only path is blocked, the robot becomes unable to move further as is violates the collision avoidance policy. Ideally, robots should be able to reason about the type of encountered obstacles (movable/unmovable). If the object is safely movable, like a small box or a trash can, as shown in Fig. \ref{fig:Thesis_illustration}, the robot should act to remove it from the path and continue executing the mission. This type of problem is known as Navigation Among Movable Obstacles (NAMO).

In this paper we propose a modular and scalable approach for overcoming movable obstacles that can work in real-time. According to this approach, a robot builds a path through a cluttered environment and interacts with movable objects to overcome them in order to continue its mission. A robot operates in an indoor environment with static obstacles and builds a 2D cost-map. We collected a dataset of 1000 synthetically created images in order to train the Convolutional Neural Network and to use it for object recognition.
We are not focused on solving the riddle-like problem, which implies calculating the order by which to move the obstacles in order to clear the path.
A significant amount of research has already been conducted on NAMO problem and related fields both in robotics and deep learning communities, such as object segmentation and recognition, semantic object representation, grounding, spatial databases for storing semantic maps, multi-layered costmap representation, reinforcement learning approach for real-world modeling dynamics and various path planning algorithms.
The proposed research and experimental work provide a further investigation into how wheeled robots can build and interactively adjust a path through a cluttered environment by manipulating the obstacles. 

\section{Related Work}
Clingerman et al. \cite{7354249} proposed an algorithm for estimating the manipulability of initially unknown obstacles. The core idea is representing the environment as an evidence grid, where each cell is associated with gamma-distributed cost and visual features of objects.

Scholz et al. \cite{7759546} proposed the first planner for the NAMO problem that can deal with object dynamics uncertainty. Their algorithm is based on hierarchical Markov-Decision Process to handle dynamics uncertainty and physics-based Reinforcement Learning framework \cite{pmlr-v32-scholz14} to ground this uncertainty into compact model space.

The most recent research on a subject was conducted by Regier et al. They proposed to use semantic information about objects for humanoid navigation through obstructed regions of the path. Their approach includes semantic segmentation and classification of the input images using CNN, assigning recognized classes and appropriate actions to the projected grid cells, path planning on a resulted costmap and execution of the object manipulation commands \cite{8625036} (Jan. 2020).

Kakiuchi et al. \cite{5650206} proposed a solution for  HRP-2 to work in unknown environment with movable obstacles with only relying on cameras and force sensor inputs. Color range sensor is used for clustering the obstacles, and active sensing is used for detecting movable obstacles.

In their studies \cite{7139972}, Scholz proposed a Physics-Based Reinforcement Learning (PBRL), and presented two approaches for modeling object dynamics: exploiting the geometric properties of physical dynamics, and by using modern physical simulation methods. 

Novin et al. in their work \cite{8593989} addressed a problem of patient falls in hospitals and developed a Patient Assistant Mobile Robot (PAM), that aims at preventing elderly people from falling, by delivering them a walker. Minimally interacting with objects, a robot uses a probabilistic method to derive the dynamic model of the legged objects.

Scholz et al. \cite{5980288} presented a navigation and manipulation approach for a humanoid robot that is pushing a cart.

For context-sensitive navigation Lu et al. \cite{6942636} proposed the improvement to use multi-layered costmaps, consisting of semantically-separated layers. 

Another approach to NAMO problem could be based on the idea proposed by Bansal et al. \cite{bansal2019combining}. In their work, they used model-based control (MPC) in combination with learning-based perception for producing series of waypoints and following them.

Miyazaki et al. \cite{8014083} proposed sweeping motions for a dual-arm humanoid robot to find an object of interest under other objects.

Numerous studies were focused on image recognition and segmentation. Long et al. contributed to cosiety \cite{shelhamer2016fully} by building "fully convolutional" networks.









\section{System Overview}

\begin{figure}[!t]
    \centering
    \includegraphics[width=8.5cm]{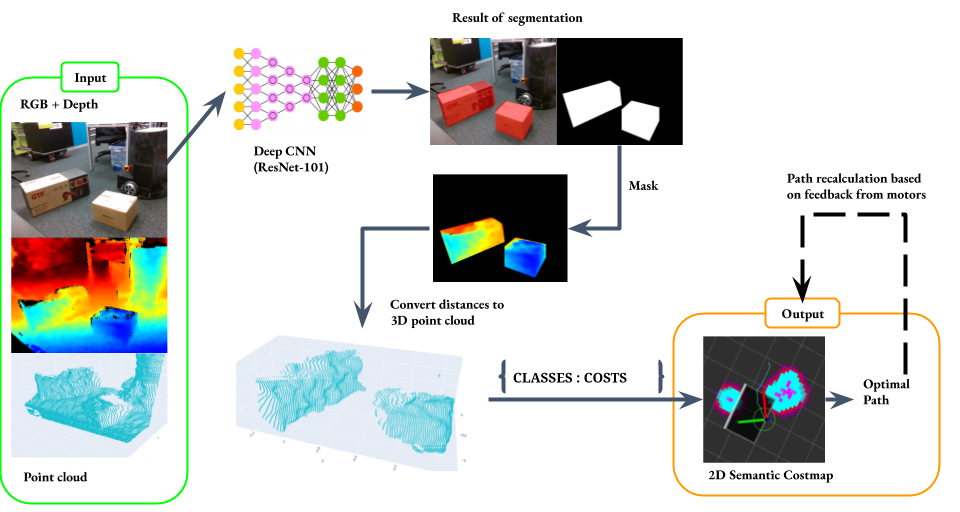}
    \caption{Overview of the proposaed approach}
    \label{fig:Overview}
    \vspace{-0.1em}
\end{figure}

In this section we describe an overview of each step and each component of our system, as illustrated in Fig. \ref{fig:Overview}.

There are four main stages: semantic segmentation of an input RGB-D image, building a 2D cost-map of obstacles, path planning on a cost-map, and path execution with feedback from motors.  Firstly, we semantically segment an RGB image and obtain a set of masks for known movable objects (boxes, trash cans), known unmovable objects (glass vases), and background. Then we align the depth image with an RGB image, apply the masks to the depth image, and transform the resulting depth pixels to an XYZ point cloud.  At this stage, we have a point cloud of recognized movable and unmovable obstacles and their labels.  After that, we project a point cloud to a 2D cost map. Points of recognized movable obstacles are assigned with estimated costs, whereas objects that are not allowed to be moved are assigned a maximum cost. After a point cloud of obstacles is built, we use an A* algorithm to find an optimal path to the next goal point, and after that, a Pure Pursuit algorithm is utilized to follow the path.  In case the robot experience difficulties with following the chosen path, for example, when a box occurs to be loaded with something heavy, the robot tries to rebuild its path.

\section{Semantic Segmentation}

\subsection{CNN Architecture}
For the image classification task, we chose to use a pre-trained Fully-Convolutional Network model with a ResNet-101 backbone \cite{shelhamer2016fully}. It was pre-trained on 20 categories that are present in the Pascal VOC dataset \cite{Everingham10}\cite{lin2014microsoft}.
We treated the FCN as a fixed feature extractor for our dataset: all weights except the final fully connected layer were frozen, and only the last layer was modified to reflect our dataset and trained after initialized with random weights.

The reason resnet-101 architecture was chosen, is that according to \cite{he2015deep} "residual connections help in handling the vanishing gradient problem", which prevents successful training of large networks. With residual connections, it is safe to train very deep layers, "because in the worst case, "unnecessary blocks" can learn to be an identity mapping and do no harm to performance."
In comparison to other architectures, for example, VGG neural network, resnet has lower complexity.

\subsection{Data Collection}
\begin{figure}[!t]
    \centering
    \includegraphics[width=8cm]{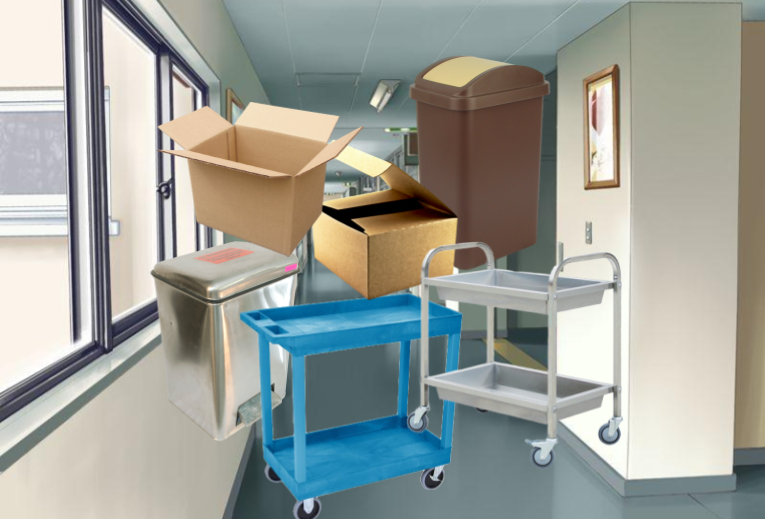}
    \caption{Synthetic dataset}
    \label{fig:dataset}
    \vspace{-0.5em}
\end{figure}

Training a deep CNN requires a large amount of annotated training images. To collect such a dataset, we wrote a script to parse images from the Internet. Using this script, we obtained 30 unique images for each of the following classes: boxes, trash cans, food trolleys, glass vases and 10 backgrounds of different environments. Each of the images either contained a transparent background or was processed using another script to remove the background.
After synthetic combination, we generated 1000 images with 1-7 randomly-placed objects on an image with known masks and labels. An example of such a synthetic image can be seen in Fig. \ref{fig:dataset}.

Since we are using transfer learning, even this minimal dataset is sufficient for generalizing.

\section{Path Planning Utilizing Obstacle Information}

In this section, we show how the newly detected objects are added to the costmap, what path-planning and path-execution algorithms are used, and how the pushing process is performed.


\subsection{Object Mapping} \label{object_mapping}

\begin{figure}[htp]
    \centering
    \includegraphics[width=8cm]{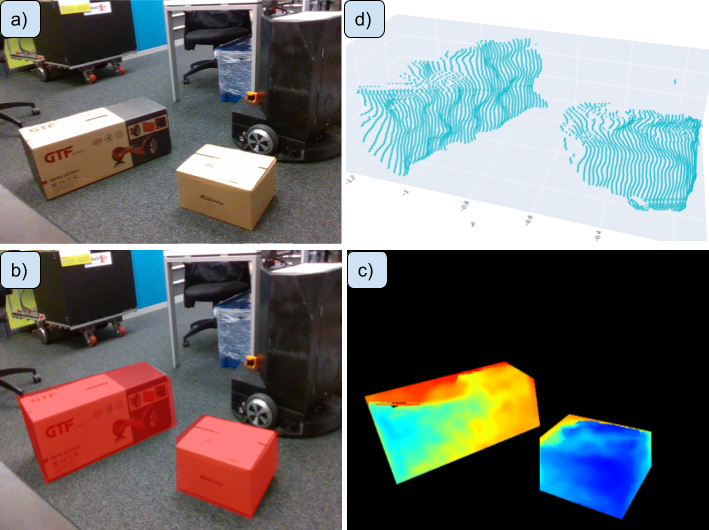}
    \caption{RGBD data processing. a) The original RGB image contains two boxes between two walls. b) Semantic segmentation results using the CNN with resnet101 backbone. c) Corresponding depth data after applying the mask d) A point cloud transformed from a depth image.}
    \label{fig:object_mapping}
    \vspace{-0.5em}
\end{figure}

This section introduces the process of obtaining point cloud coordinates from the detected and classified obstacles. First, an RGB image (Fig. \ref{fig:object_mapping}a)  is processed with a CNN to detect the known classed of obstacles and get the mask (Fig. \ref{fig:object_mapping}b). Then, the obtained mask is applied to the depth image (Fig. \ref{fig:object_mapping}c) and the resulting set of depth pixels are transformed to a point cloud (Fig. \ref{fig:object_mapping}d).

\subsection{Costmap}

Each detected and classified movable obstacle gets represented as a XYZ point cloud and after post-processing is added to a global map. Post-processing includes removing the outliers with a Statistical Outlier Removal filter \cite{BALTA2018348}, reducing the number of points and projecting the resulting set of points into a 2D cost gridmap. Each of the resulting cells is associated with object ID, the corresponding cost and projected cells on a costmap.
The cost of moving for each object is an heuristic value, associated with approximate time needed for the robot to remove the object from the way.
All other unknown occupied cells, as well as classified unmovable obstacles, are marked as fatal and robot is not allowed to collide with them. Then, the obstacles are uniformly inflated to a pre-defined radius.


\subsection{Path Planning}
The Navigation Among Movable Obstacles (NAMO) problem is similar to regular navigation task with the main difference, that the path can be occluded by movable obstacles, and the robot can attempt to move them away. The location of objects in the environment is not specified in advance and robot constantly adds the newly detected and classified obstacles on the costmap. The global planner as an input receives the goal point, the costmap with detected obstacles, both movable and static, and the position of the robot from localization system. For path planning the Generic Robot Navigation (GeRoNa) framework \cite{Huskic2018GeRoNaGR} is used. For localization and mapping the robot runs a Cartographer simultaneous localization and mapping (SLAM) algorithm, presented in \cite{45466}. The planner is based on an A* algorithm \cite{4082128}. It builds a plan for the robot to reach the goal. The global planner doesn't consider the kinematics of the robot and treats it as a circular object of a defined radius.

\subsection{Path Execution}

In this section we describe the process of path following and pushing approach.
The robot is capable of executing two primitives: move forward and rotate in place; also, it can rotate while moving forward. Path execution is based on a Pure Pursuit algorithm, which is also a part of a GeRoNa framework. When the robot detects a movable obstacle that can't be crossed around in a collision-free manner, the robot needs to evaluate the situation and choose the minimally invasive action to perform to continue its mission. The robot slowly starts to push the obstacle either until the path is opened enough for the planner to recalculate the feasible route, or until the motor current limit is reached.
This type of action can be useful in a scenario when a robot is artificially trapped, for example, with three ice-cream boxes by kids in a store.

\section{Experimental Evaluation}
In this section, we present the evaluation results of our classification neural network, and we show the performance of our planner in a simulation with different setups.





\subsection{Classification Results}
\begin{figure}[!t]
    \centering
    \includegraphics[width=8cm]{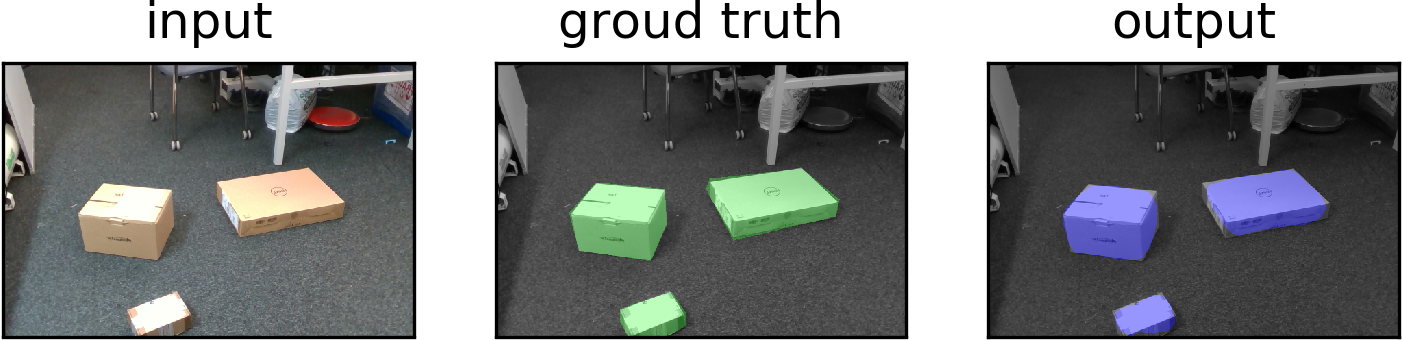}
    \includegraphics[width=8cm]{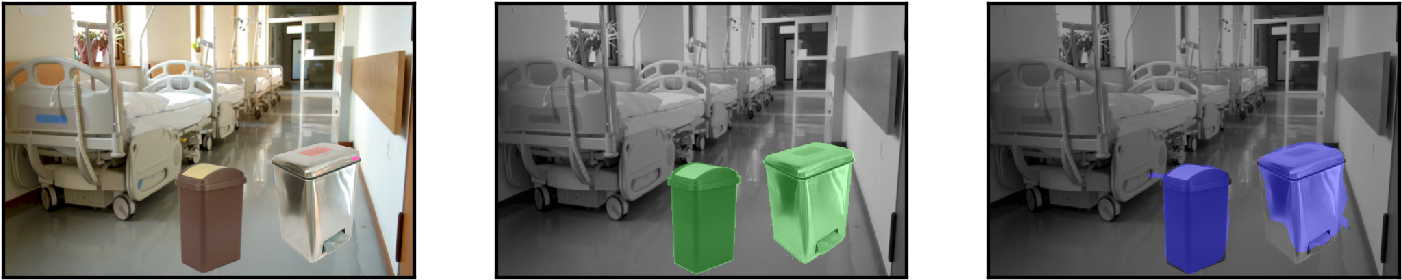}
    \caption{Segmentation results}
    \label{fig:segm_result}
    \vspace{-0.5em}
\end{figure}

The segmentation part was implemented using Pytorch. We have trained models on a custom dataset with on-the-fly data augmentation, including random intensity shifts, rotation, translation, and scaling.
For a resolution of 640 x 480 the resulted IoU is 0.904 for boxes and 0.879 for trash cans.
FCN ResNet-101 and DeepLabv3 ResNet-101 models were trained for 100 epochs using an NVIDIA Tesla V100 16GB GPU. For supervised learning, we use the SGD optimizer with a learning rate of 0.001 and cross-entropy loss weighted by the class cardinality to avoid class imbalance. Segmentation results were evaluated in terms of intersection over union (IoU) and cross-entropy (CE) metrics to estimate both a pixel-wise and overall quality of segmentation outputs. The segmentation results for different classes can be seen in Fig \ref{fig:segm_result}. 

Table \ref{table:table_segmentation} shows performance of models for segmentation of boxes and trash cans. It justifies the selection of FCN ResNet-101 architecture, which performs best for the segmentation of boxes in terms of all chosen metrics.

Table \ref{table:table_segmentation} shows performance of models for segmentation of boxes and trash cans. It justifies selection of FCN ResNet-101 architecture, which perform best for segmentation of boxes in terms of all chosen metrics.

\begin{table}[ht]
\caption{Segmentation Results.}
\label{table:table_segmentation}
\begin{tabular}{|l||l l| l l|}
\hline
\multirow{2}{*}{Approach} & \multicolumn{2}{c|}{Boxes} & \multicolumn{2}{c|}{Trash cans} \\
\cline{2-5} 
& IoU,\% & CE & IoU,\% & CE \\ \hline \hline
FCN ResNet-101 & 90.4 & 0.066 & 74.1 & 0.094 \\ \hline
DeepLabv3 ResNet-101 & 87.9 & 0.074 & 80.4 & 0.125 \\ \hline
\end{tabular}
\end{table}

\subsection{Simulation Experiments}
\begin{figure}[htp]
    \centering
    \includegraphics[width=8cm]{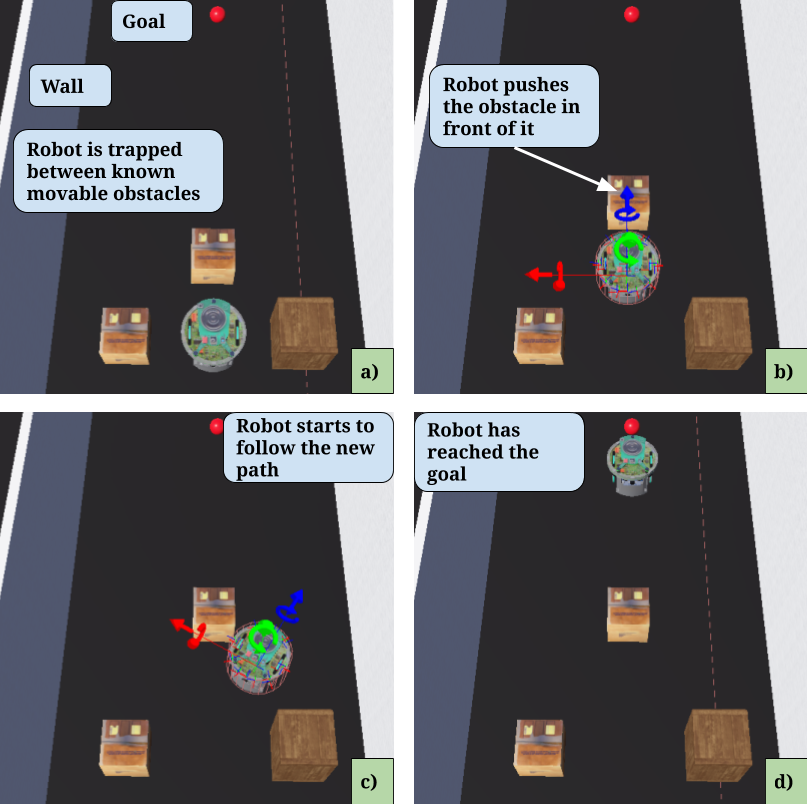}
    \caption{Scenario 1. The red dot indicates a goal. a) The robot is trapped between card boxes. b) The robot chooses to push the box in front until the feasible path is recalculated. c) The robot starts to follow the new path. d) The robot has reached its goal.}
    \label{fig:scenario_0}
    \vspace{-0.5em}
\end{figure}

\begin{figure}[htp]
    \centering
    \includegraphics[width=8cm]{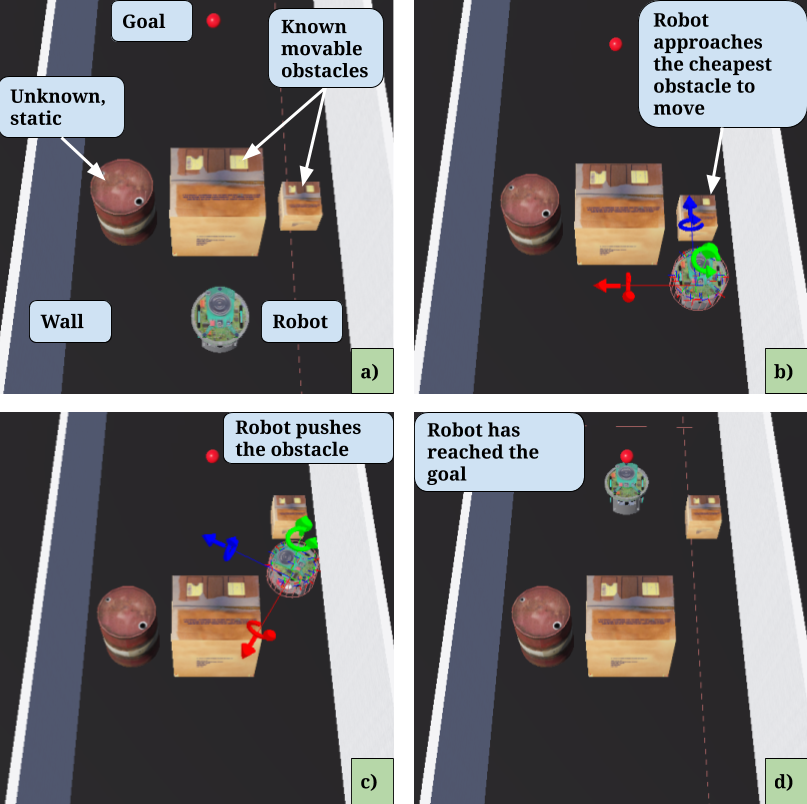}
    \caption{Scenario 2. The red dot indicates a goal. a) Starting position. Three obstacles are blocking the path. b) The robot chooses to push the lightest obstacles among known - a box on the right c) The robot pushes the box until the new feasible path is calculated. d) The robot has reached its goal.}
    \label{fig:scenario_1}
    \vspace{-0.5em}
\end{figure}

For simulation, we consider a simplified environment where the robot operates in a narrow corridor and the chosen set of obstacles are blocking the path in one of the following configurations.
In the first scenario the robot is trapped inside three icecream boxes, as shown in Fig. \ref{fig:scenario_0}.
The second scenario is shown in Fig. \ref{fig:scenario_1}, where the path is blocked by different types of objects, and the robot searches for the cheapest path to follow.
The red dot indicates the goal and different boxes and oil barrel represent the obstacles.


\subsection{Real-World Experiments}

The real-world experiments are conducted on a differential drive robot of circular form. All high-level algorithms are built as ROS nodes and are running on an Intel NUC computer in real-time. The robot is equipped with a Hokuyo 2D LIDAR and an Intel Realsense RGB-D camera. Motors are controlled with an STM-32f4 microcontroller. The Realsense camera is oriented vertically and inclined by 30$^{\circ}$ for a larger field of view on the front.

\section{Conclusions and Future Work}
\subsection{Conclusions}
In this work we have proposed a modular approach for the wheeled robots to deal with Navigation Among Movable Obstacles (NAMO) problem. The task is to find an optimal path to the target position, that may be occluded by movable obstacles while avoiding collisions with static obstacles. According to this approach, the robot employs a convolutional neural network (CNN) to detect and classify which of the obstacles are movable and which are not, and builds a 2D costmap of the environment. For the robot to overcome the cluttered area, a path planning algorithm calculates a path on a costmap by exploiting the knowledge about classes. In case where no clear path is available, the robot needs to perform active sensing of encountered movable obstacle in order to understand its physical parameters, such as mass and friction. 
The robot combines knowledge of class produced by CNN with a current level measured on wheels to choose appropriate action. The robot can push an obstacle away from the path in a chosen direction, or recalculate the path and perform another attempt. 
The multi-layered costmap is being updated continuously during navigation and includes static obstacles, classified movable obstacles, classified unmovable obstacles and costs of appropriate actions.
Our trained neural network proved to be capable of distinguishing boxes, food trolleys, trash cans and glass vases and can be extended to many other types of objects. 
The various experiments in simulation showed that the robot succeeded in getting to the goal through a cluttered region by pushing a movable obstacle out of the way or going around the obstacles without collision.

\subsection{Future work}
There are several challenges left unsolved that would be addressed in further research, as well as real-world experiments. The robot should be able to safely move encountered obstacles while accounting for the ones that are fragile or valuable and are not supposed to be touched. For that, the robot needs to have a parameterized representation of the physical dynamics of the objects in terms of probability distributions. In addition, we plan to improve object recognition performance by merging classification predictions from both sources - RGB data and depth measurements. Another vital ability is reasoning about an order, in which obstacles should be moved in order to clear the path. Measuring the current level on the wheels should be replaced with force sensors due to their universal measurements and their independence from robot setup.
The cost function of motion can also account for the power consumption or risk of failure.

The developed approach could be used for enhancing a robot's performance in such industrial tasks as autonomous delivery, cleaning, disinfection and stocktaking. 


\addtolength{\textheight}{-12cm}   


\bibliographystyle{unsrt} 
\bibliography{bib}

\end{document}